\documentclass{article} 
\usepackage{iclr2020_conference,times}

\usepackage{hyperref}
\usepackage{url}
\usepackage{amsmath}
\usepackage{graphicx}
\usepackage{subcaption}
\usepackage{textcomp}

\title{Towards a predictive spatio-temporal representation of brain data}

\author{Tiago Azevedo \\
Department of Computer Science and Technology\\
University of Cambridge\\
\texttt{tiago.azevedo@cst.cam.ac.uk} \\
\And
Luca Passamonti \\
Department of Clinical Neurosciences \\
University of Cambridge \\
\texttt{lp337@medschl.cam.ac.uk} \\
\AND
Pietro Li\`{o} \\
Department of Computer Science and Technology\\
University of Cambridge\\
\texttt{pietro.lio@cst.cam.ac.uk}
\AND
Nicola Toschi \\
Department of Biomedicine and Prevention \\
University of Rome ``Tor Vergata'' \\ \\
A. A. Martinos Center for Biomedical Imaging \\
Massachusetts General Hospital and Harvard Medical School \\
\texttt{toschi@med.uniroma2.it}
}

\iclrfinalcopy 
\begin{document}

\maketitle
%
%

\begin{abstract}
The characterisation of the brain as a ``connectome'', in which the connections are represented by correlational values across timeseries and as summary measures derived from graph theory analyses, has been very popular in the last years. However, although this representation has advanced our understanding of the brain function, it may represent an oversimplified model. This is because the typical fMRI datasets are constituted by complex and highly heterogeneous timeseries that vary across space (i.e., location of brain regions). We compare various modelling techniques from deep learning and geometric deep learning to pave the way for future research in effectively leveraging the rich spatial and temporal domains of typical fMRI datasets, as well as of other similar datasets. As a proof-of-concept, we compare our approaches in the homogeneous and publicly available Human Connectome Project (HCP) dataset on a supervised binary classification task. We hope that our methodological advances relative to previous ``connectomic'' measures can ultimately be clinically and computationally relevant by leading to a more nuanced understanding of the brain dynamics in health and disease. Such understanding of the brain can fundamentally reduce the constant specialised clinical expertise in order to accurately understand brain variability.
\end{abstract}

\section{Introduction}

The study of functional Magnetic Resonance Imaging (fMRI) is particularly important to healthcare as it has been bringing insights on many psychological constructs like motivation, attention, working memory, and others. These constructs are, in turn, very close to the symptoms and behavioural problems that many patients with various brain disorders experience everyday (e.g., apathy, impulsivity, and motor deficits). Although brain anatomical differences underlie brain function (i.e., they are the ``substrate'' of the brain function), it is the brain functionality within specific circuits that will determine how we all behave and function at the mental/psychological level. One considerable issue with fMRI and similar data (e.g., EEG/MEG scans) is that their timeseries are noisy and volatile, making it a very challenging type of data to analyse. Making simplifications on this data representation is then very helpful to easily remove the inherent complexity and highly dynamic information.

The fMRI timeseries are typically studied with correlational matrices and graph theory techniques. Although these approaches have significantly deepened our understanding of the brain function, they have also been limited by the oversimplification of the fMRI timeseries in terms of their spatial (i.e., brain location) and temporal architecture (i.e., timeseries). We believe that the last developments in Machine Learning (ML) and, in particular, Geometric Deep Learning (GDL), were not yet fully exploited in the analyses of fMRI datasets. Given the particularly high-dimensional nature of fMRI timeseries, more specific models that can leverage on the richness of such structure might improve our understanding of brain function in health and disease.

The aim of this paper is then to formulate and compare novel architectures based on GDL and convolutional operators to combine representations from the two components of fMRI data - spatial and temporal - into a single, deep learning model which can be trained for any prediction task. As a proof-of-concept, we will use the homogeneous and publicly available Human Connectome Project (HCP) dataset to predict binary sex. We hope our various comparisons and developed models can contribute to laying the groundwork for explicitly incorporating all the information from fMRI scans.

\section{Related Work}

Our models take inspiration from the field of GDL to account for the spatial interrelationships between brain regions, and uses convolution operators to account for the temporal dynamics of fMRI data. The field of GDL has witnessed fast developments in recent years~\citep{zhou2018graph, wu2019comprehensive}, making GDL architectures the natural modelling approach when analysing graphs. Graph Convolutional Networks (GCN) have been applied on brain connectivity data for sex classification~\citep{Arslan2018}, and Convolutional Neural Networks (CNN) have been successfully applied to the analysis of timeseries~\citep{Liu2019}. Given the challenges inherent to fMRI data (e.g., sample size and noise), the number of DL applications is not as great as when looking for graph theory analysis; however, a number of studies have applied DL to fMRI data, including the classification of brain disorders using Siamese-inspired neural network~\citep{Riaz2020}, but without GDL layers.

A few papers have presented methods that leverage both spatial and temporal components in various domains such as skeleton-based action recognition~\citep{yan2018spatial}, and traffic forecasting~\citep{diao2019dynamic}. These approaches usually define a single spatio-temporal convolutional block in which, for each timestep, one operation is performed across the graph. While such approaches are well suited for these specific problems (e.g., creation of meaningful embeddings on each timestep), they are conceptually less compatible with the analysis of brain data for a final prediction task.

\section{Proposed Model}

\subsection{Modular Architecture}
\begin{figure}[h!]
\centering
\includegraphics[width=.9\linewidth]{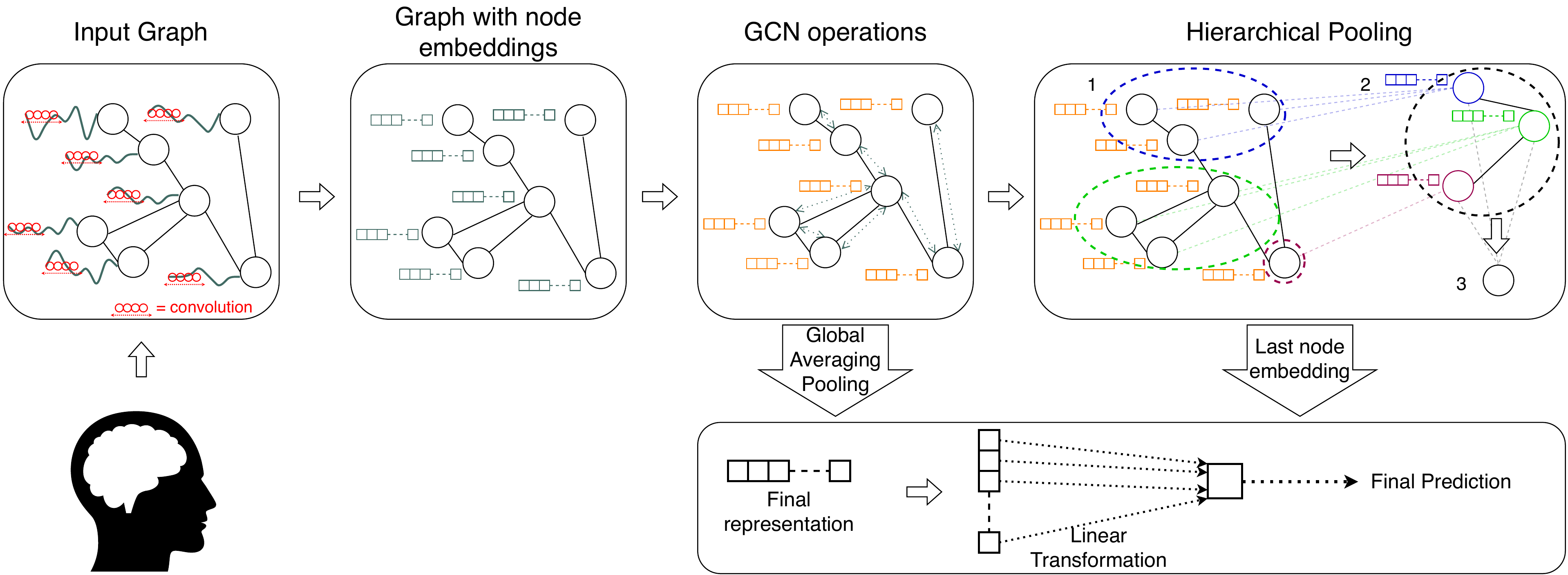}
\caption{Overview of a spatio-temporal model. The first step creates embeddings for each node through 1D convolution operators, followed by a GCN layer to transform each node's features according to information passed from its neighbours. A final representation of the graph can be created either using a global averaging pooling, or using a hierarchical pooling. A final set of linear transformations is then applied to generate the final prediction}
\label{fig:st_model}
\end{figure}

Our models were developed using Pytorch~\citep{NEURIPS2019_9015} and, for the GDL parts, in Pytorch Geometric~\citep{FeyLenssen2019}. Our architectures expect a graph-structured data as input, in which each node (i.e. brain region) corresponds to a single timeseries (i.e. the fMRI timeseries sampled in that brain region). The data for each sample consists of (1) a feature matrix $\mathbf{X}$ of size $N \times T$, where $N$ is the number of nodes, and $T$ the length of the timeseries (cf. Section~\ref{sec:experiments}), and (2) a sparse or dense representation of the adjacency matrix. The sparse representation will have size of $2 \times E$, where $E$ is the number of edges, while the dense representation will be the adjacency matrix $\mathbf{A}$ itself, with size $N \times N$. Given that fMRI data can be noisy and prone to outliers, we normalised each timeseries through Robust Scaling. The overall architecture is summarised in Figure~\ref{fig:st_model}. 

\subsection{Temporal Feature Extraction}

We use 1D convolutional operations to create feature representations from the temporal dynamics in each node. We employed a CNN with 4 layers of \textit{1D Convolution}, \textit{1D Batch Normalisation}, and \textit{relu activation}, whose hyperparameters can be seen in Appendix~\ref{app:hyperparam_info}. After these 4 layers, the features from all channels are flattened and passed through a linear transformation to reduce the representation to a fixed embedding of size 256. Overall, this compresses the original $N \times T$ feature matrix into an activation matrix $\mathbf{H}^{(1)}$ of size $N \times 256$.

We have also used strided Temporal Convolutional Networks (TCN), originally described by~\cite{bai2018empirical}. This specific type of CNN is built on top of normal convolutional networks, but with increasing values of dilation as more layers are used. They are also considered causal convolutions because an output at a specific timestep is only convolved with elements from that timestep and earlier elements from the previous layer. We used weight normalisation, as well as a skip connection in each block, but to make it comparable with the CNN, the remaining hyperparameters are the same.

\subsection{GCNs for Spatial Information Sharing}

We explicitly allow for message passing (i.e. information exchange) across nodes. This message-passing mechanism will update our previous activation matrix $\mathbf{H}^{(1)}$ to a new activation matrix, $\mathbf{H}^{(2)}$. This preserves matrix size but updates each node with a share of information pulled from its neighbours. Effectively, this means that an operation is applied which is defined as a function of the previous activation matrix and of the adjacency matrix: $f\left ( \mathbf{H}^{(1)}, \mathbf{A} \right )$. In our architecture, this function $f$ is implemented through the graph convolutional operator defined by~\cite{kipf2016semi} as follows:

$$
\mathbf{H}^{(2)} = \sigma\left( 
 \mathbf{\tilde{D}}^{-1/2} \mathbf{\tilde{A}} \mathbf{\tilde{D}}^{-1/2} \mathbf{H}^{(1)} \mathbf{W}\right ),
$$

where $\mathbf{\tilde{A}} = \mathbf{A} + \mathbf{I}$ ($\mathbf{I}$ is the identity matrix), $\mathbf{\tilde{D}}$ is the diagonal node degree matrix of $\mathbf{\tilde{A}}$, and $\mathbf{W}$ is a learnable weight matrix on the layer where the operator is applied.

\subsection{Pooling Mechanisms for downstream prediction task}

The activation matrix $\mathbf{H}^{(2)}$ needs to be pooled in order to be employed in a final downstream prediction task. The simplest option is the adoption of a global mean pooling mechanism that averages the activation matrix for each feature dimension across all nodes. This transforms the activation matrix with size $N \times 256$ to a final representation with size $1 \times 256$.

Under the assumption of locally-dependent degrees of importance (for the prediction task itself) across distinct nodes (i.e. brain regions), a hierarchical pooling representation might create richer embeddings. In other words, it might be desirable to account for different importances of brain regions according to the specific prediction task. We therefore implemented and included the option to employ the differentiable pooling operator introduced by~\cite{ying2018hierarchical}, in which the operator learns how to sequentially map nodes to clusters on the graph until the generation of a final single node with the hierarchical representation. This differentiable pooling mechanism, at the $l$-th layer, can be defined as a function called $\mathrm{DiffPool}$ which receives an adjacency matrix and node embedding matrix, and returns a new adjacency matrix as well as new node embeddings:

$$
\mathbf{A}^{(l+1)}, \mathbf{X}^{(l+1)} = \mathrm{DiffPool}\left(\mathbf{A}^{(l)}, \mathbf{X}^{(l)} \right)
$$

This is achieved through the application of a graph neural network. Specifically, in $\mathrm{DiffPool}$ the same graph neural network architecture is duplicated and applied separately to create a new embedding $\mathbf{Z}$, as well as an assignment matrix $\mathbf{S}$ as follows:

$$
\mathbf{Z}^{(l)} = \mathrm{GNN}_{l, \mathrm{embed}} \left( \mathbf{A}^{(l)}, \mathbf{X}^{(l)} \right) \quad \text{and}\quad \mathbf{S}^{(l)} = \mathrm{softmax} \left( \mathrm{GNN}_{l, \mathrm{pool}} \left( \mathbf{A}^{(l)}, \mathbf{X}^{(l)} \right) \right),
$$

where $\mathbf{S}^{(l)}$ will have size $n_{(l)} \times n_{(l+1)}$, $n_{(l)}$ is the number of nodes in the $l$-th layer, and $n_{(l+1)}$ the (new) number of clusters in the following layer (i.e., 25\% of the previous nodes). As recommended by the original paper~\citep{ying2018hierarchical}, we employed 3 layers of $\mathrm{GraphSAGE}$~\citep{hamilton2017inductive} followed by a 1D batch normalisation, with a final skip connection. With these definitions, and as outlined in the original paper, we implemented two layers of $\mathrm{DiffPool}$ as described in the following two equations:

$$
\mathbf{X}^{(l+1)} = {\mathbf{S}^{(l)}}^T \mathbf{Z}^{(l)} \quad \text{and}\quad \mathbf{A}^{(l+1)} = {\mathbf{S}^{(l)}}^T \mathbf{A}^{(l)} \mathbf{S}^{(l)},
$$

where $\mathbf{X}^{(l+1)}$ has size $n_{(l+1)} \times F$ (where $F$ is the number of features), and $\mathbf{A}^{(l+1)}$ has size $n_{(l+1)} \times n_{(l+1)}$.

\section{Experiments}
\label{sec:experiments}

\subsection{Dataset Description}

We employed the \textbf{Human Connectome Project} (HCP) fMRI Data, which is made available as part of the S1200 PTN release\footnote{\url{https://www.humanconnectome.org/study/hcp-young-adult/document/extensively-processed-fmri-data-documentation}}. 4 fMRI sessions of 15-minute multi-band acquisition (TR = 0.72s) were acquired and preprocessed by HCP for 1003 healthy subjects. A group-wise spatial independent component analysis (ICA) was run by HCP to obtain spatio-temporal decompositions at different dimensionalities, from which subject- and component- specific time series were extracted by dual regression. In this paper we employ the data projected onto 50 distinct spatio-temporal components, where each component represents one node. 

\subsection{Experimental setup}

For each subject, we employed the four distinct sessions/samples per subject with 1,200 timesteps for each sample and component. We also did a further divisions where only 75 timesteps (approximately 1 minute of scan time) were considered per node, meaning 64 samples per person. We randomly downsampled the dominant class for balancing purposes, resulting in a dataset of 3,752 and 60,032 samples, respectively. Edge connections were defined as full correlations using the Ledoit Wolf covariate estimator available in the nilearn python package\footnote{\url{https://nilearn.github.io/}}. Our results are reported using a Stratified Group 5-fold cross validation with hyperparameter search (more details in Appendix~\ref{app:hyperparam_info}). We explored two thresholding strategies for the adjacency matrices: 5\% and 20\% thresholds in strength followed by binarisation. For comparison, we also report results using a XGBoost model on the correlation values only (cf. Appendix~\ref{app:hyperparam_info}). For ablation analysis, we provide results with and without a GCN layer.

%
%

\subsection{Results}

Table~\ref{tab:results} summarises the results of the experiments. A further illustration of our results using the best spatio-temporal model (i.e. \textbf{mean\_CNN\_GCN5}) is illustrated in Appendix~\ref{app:roc_curve}.

\begin{table}[h]
\centering
\caption{Summary of results reported as means over a 5-fold CV procedure and standard deviation in parenthesis. In bold the best results for the XGBoost, and spatio-temporal models when standard deviation is below 0.1. The model name conventions are: \textbf{mean} when using global averaging pooling; \textbf{diff} when using hierarchical pooling, with a number representing the threshold; \textbf{CNN} and \textbf{TCN} when using the respective convolutional network on the temporal part; \textbf{GCN} when a GCN layer is used; \textbf{4split}/\textbf{64split} when using 4/64 samples per person, respectively; \textbf{bin} when using binarisation at 5\% on the flatten array}
\label{tab:results}
\begin{tabular}{llllr}
\multicolumn{1}{c}{\textbf{Model}} & \multicolumn{1}{c}{\textbf{ AUC }} & \multicolumn{1}{c}{\textbf{ Sensitivity }} & \multicolumn{1}{c}{\textbf{ Specificity }} & \textbf{\# Parameters}  \\ 
\hline
diff5\_CNN                         & 0.572 ( 0.078 )                    & 0.722 ( 0.372 )                            & 0.375 ( 0.369 )                            & 2713886                 \\
diff20\_CNN                        & 0.605 ( 0.081 )                    & 0.638 ( 0.340 )                            & 0.507 ( 0.320 )                            & 2713886                 \\
diff5\_CNN\_GCN                    & 0.534 ( 0.063 )                    & 0.799 ( 0.400 )                            & 0.203 ( 0.398 )                            & 2779678                 \\
diff20\_CNN\_GCN                   & 0.526 ( 0.105 )                    & 0.657 ( 0.354 )                            & 0.428 ( 0.373 )                            & 2779678                 \\ 
\hline
mean\_TCN                          & 0.674 ( 0.046 )                    & 0.558 ( 0.330 )                            & 0.614 ( 0.340 )                            & 1249545                 \\
mean\_TCN\_GCN5                    & 0.649 ( 0.046 )                    & 0.771 ( 0.390 )                            & 0.276 ( 0.390 )                            & 1315337                 \\
mean\_CNN                          & 0.689 ( 0.035 )                    & 0.626 ( 0.075 )                            & 0.622 ( 0.047 )                            & 1248545                 \\
mean\_CNN\_GCN5                    & \textbf{0.701 ( 0.037 )}           & \textbf{0.645 ( 0.065 )}                   & 0.635 ( 0.036 )                            & 1314337                 \\
mean\_CNN\_GCN20                   & 0.700 ( 0.038 )                    & 0.636 ( 0.067 )                            & \textbf{0.641 ( 0.029 )}                   & 1314337                 \\ 
mean\_CNN\_64split                 & 0.646 ( 0.023 )                    & 0.620 ( 0.057 )                            & 0.587 ( 0.056 )                            & \textbf{101665}                  \\
\hline
xgboost\_bin\_4split               & 0.688 ( 0.024 )                    &         0.700 ( 0.027 )                    &         0.676 ( 0.031 )                    &  \multicolumn{1}{c}{-}      \\
xgboost\_4split                    & \textbf{0.785 ( 0.007 )}           & \textbf{0.784 ( 0.017 )}                   & \textbf{0.787 ( 0.019 )}                   &  \multicolumn{1}{c}{-}    \\
xgboost\_64split                   & 0.690 ( 0.006 )                    & 0.678 ( 0.006 )                            & 0.701 ( 0.010 )                         &     \multicolumn{1}{c}{-}    \\
\hline
\end{tabular}
\end{table}

\section{Discussion and Conclusions}

Overall performance is comparable with previous papers dealing with sex prediction from fMRI data~\citep{Weis2019}. In this implementation, both TCNs and hierarchical pooling mechanisms did not provide advantages when compared to CNN and global averaging pooling, respectively. This can be seen as the high sensitivity is accompanied by a higher deviation on test folds, while also increasing the number of parameters to learn. Contrary to previous research on graph classification tasks, using a GCN layer did not beat the more established methods that only take a flatten representation into account (i.e. XGBoost). There is then a seemingly incompatibility between GDL models and temporal fMRI data which we think has not been addressed adequately so far. This likely indicates that the importance of all brain regions is fairly balanced for this particular prediction task; however, this assumption needs further exploration, especially in terms of data volume, as it is possible that the more complex spatial models may be too complex to be trained on this particular dataset. One interesting result that needs to be considered for future work, is the one when considering only binarised values over a threshold of 5\% (i.e. xgboost\_bin\_4split model), which shows that the spatial information of these graphs alone contains already sufficient information to achieve considerably good metrics. Given the superiority of global averaging pooling, we further hypothesise that the GCN layer could not achieve a clear improvement in performance as maybe nodes do not need to learn any further spatial structure as an averaged representation over all nodes is already enough for this prediction task. This could explain the inconsistent results between the two thresholding strategies, but additional analysis is needed for other brain spatial structures. 

Although the XGBoost model on the flatten correlation values gives the best metrics, our results are a promising first step towards a stronger spatio-temporal representation of the brain. A clear next step can be inferred from the results when using only 75 timepoints per brain region instead of 1,200 (i.e., the \textit{\_64split} models): the metrics are comparable even though the dimensionality was heavily reduced. We conjecture that the high level of noise in such timeseries makes it difficult for the convolutional operators to learn meaningful representations, hence learning similar embeddings regardless of the amount of timepoints considered. So, other methods are needed to capture richer node embeddings (e.g., autoencoders or convolutions over the spatial domain). This is even more compelling as there is current research analysing how the brain connectivity changes throughout a single fMRI scan, instead of having a stationary behaviour~\citep{Preti2017}. 

In conclusion, our architectures can be extendable to include, for example, any connectomic type of data (e.g., structural, multimodal, or gene co-expression). This could ultimately bring meaningful insights and personalised therapies that might emerge from the study of these connectomes, as for example the impact of drug responses and how pain variability is encoded in brain communication~\citep{Kucyi2015}. We feel that this comparative study can contribute to laying the groundwork for explicitly incorporating spatio-temporal information into every prediction task in neuroscience.


\subsubsection*{Acknowledgments}
Data were provided by the Human Connectome Project, WU-Minn Consortium (PIs: David Van Essen and Kamil Ugurbil; 1U54MH091657) funded by the 16 NIH Institutes and Centers that support the NIH Blueprint for Neuroscience Research; and by the McDonnell Center for Systems Neuroscience at Washington University. T Azevedo is funded by the W. D. Armstrong Trust Fund, University of Cambridge, UK. L Passamonti is funded by the Medical Research Council grant (MR/P01271X/1) at the University of Cambridge, UK. The Titan V GPUs employed in this research were generously donated to N Toschi by NVIDIA.

\bibliography{iclr2020_conference}
\bibliographystyle{iclr2020_conference}

\newpage

\appendix
\section{Hyperparameter Information}
\label{app:hyperparam_info}
For the 1D convolutional layers, the kernel size is 7 (with weights initialised by sampling from a normal distribution $\mathcal{N}\left( 0, 0.01^2 \right)$), and the number of channels used by the convolutions are, in the following order: 1, 8, 16, 32, and 64. By using a padding of 3 and a stride of 2, the length of the timeseries is reduced by a factor 2 after each layer. 

Our results are reported using a Stratified Group 5-fold cross validation within a train/validation/test setup. Within each training fold, one fifth of the data was employed as a validation set for hyperparameter search. Both the outer 5 folds, as well as the inner division between training/validation sets, account for the sex label and the information of the number of the scan (depending on the experiment each person could have been splitted in 4 scans or 64 scans) for balanced stratification purposes. Additionally, given that we employed repeated measures for each subject, we forced that no data from the same subject could be present in different folds at the same time to avoid data leakage. The hyperparameter search consisted of a grid of values of dropout (0, 0.5, and 0.7), learning rate (1e-4, 1e-5, and 1e-6), and weight decay (0.005, 0.5, and 0), in which the model with the lowest loss in the validation set was selected as the best model to be evaluated in the test set. We used a batch size of 500 over 30 epochs for all cases, except when using TCN (batch size of 400), and the split in 64 samples per person (batch size of 1000); all using the Adam optimiser. 

When using the XGBoost model, the same cross validation procedure is applied. The hyperparameter search, in this case, consisted of a grid of values of min\_child\_weight, gamma, subsample, colsample\_bytree, and max\_depth.

\section{Receiver Operating Characteristic on one Model}
\label{app:roc_curve}

\begin{figure}[h!]
\centering
\includegraphics[width=.7\linewidth]{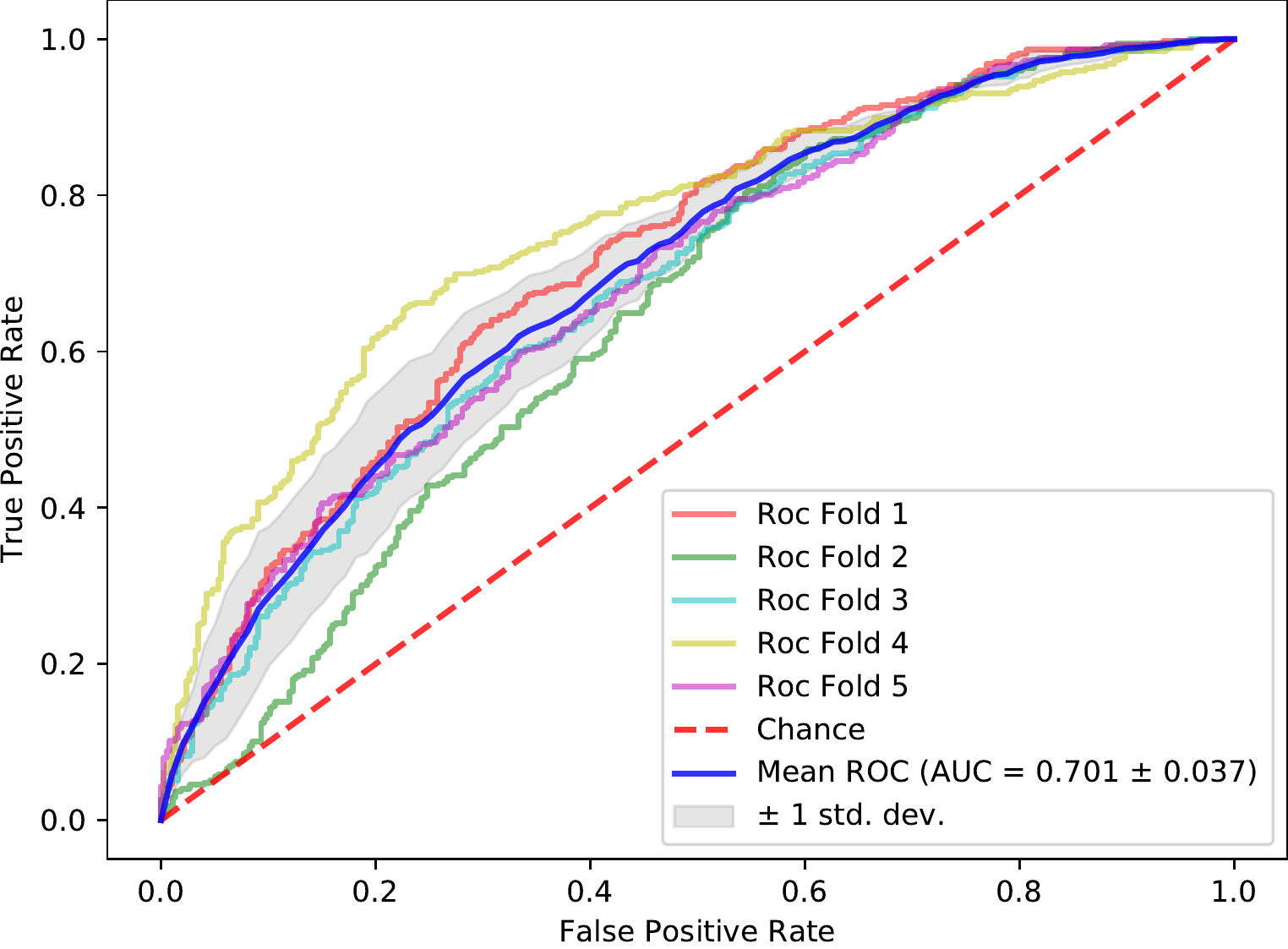}
\caption{Receiver Operating Characteristic curves for the model using global averaging pooling with a CNN on the temporal part, and a GCN layer on a thresholded (5\%) adjacency matrix}
\end{figure}

\end{document}